\documentclass{article}
\usepackage{spconf,amsmath,epsfig}
\usepackage{graphicx}
\usepackage{amsfonts,amssymb}
\usepackage{multirow}
\usepackage{float}
\usepackage{color}
\usepackage{makecell}

\let\OLDthebibliography\thebibliography
\renewcommand\thebibliography[1]{
  \OLDthebibliography{#1}
  \setlength{\parskip}{0pt}
  \setlength{\itemsep}{0pt plus 0.3ex}
}

\begin{document}\sloppy

\def\x{{\mathbf x}}
\def\L{{\cal L}}

\title{Reference-Aided Part-Aligned Feature Disentangling\\ for Video Person Re-Identification}
%
\name{Guoqing Zhang, Yuhao Chen, Yang Dai, Yuhui Zheng, Yi Wu}
\address{}

\maketitle

\begin{abstract}
Recently, video-based person re-identification (re-ID) has drawn increasing attention in compute vision community because of its practical application prospects. Due to the inaccurate person detections
and pose changes, pedestrian misalignment significantly increases the difficulty of feature extraction and matching. To address this problem, in this paper, we propose a \textbf{R}eference-\textbf{A}ided \textbf{P}art-\textbf{A}ligned (\textbf{RAPA}) framework to disentangle robust features of different parts. Firstly, in order to obtain better references between different videos, a pose-based reference feature learning module is introduced. Secondly, an effective relation-based part feature disentangling module is explored to align frames within each video. By means of using both modules, the informative parts of pedestrian in videos are well aligned and more discriminative feature representation is generated. Comprehensive experiments on three widely-used benchmarks, i.e. iLIDS-VID, PRID-2011 and MARS datasets verify the effectiveness of the proposed framework. 
Our code will be made publicly available.
\end{abstract}

\section{Introduction}
Person re-identification (re-ID) is an important retrieval task to match pedestrian images or videos captured from multiple non-overlapping cameras. Because of its wide application prospects in public safety and video surveillance, person re-ID has attracted increasing interest in recent years. Due to complicated and variable visual variations in practical scenarios such as pose, viewpoints, occlusion, illumination and background clutter, it remains a challenging task.

Currently, great progress has been made in image-based person re-ID \cite{1, 2, 3}, and video-based person re-ID has drawn increasing attention because of the impressive benefits of using multiple images, which can provide tremendous temporal information \cite{4, 5}. In this paper, we focus on the person re-ID problem in the video setting.

\begin{figure}[t]
\centering
\includegraphics[width=0.8\columnwidth]{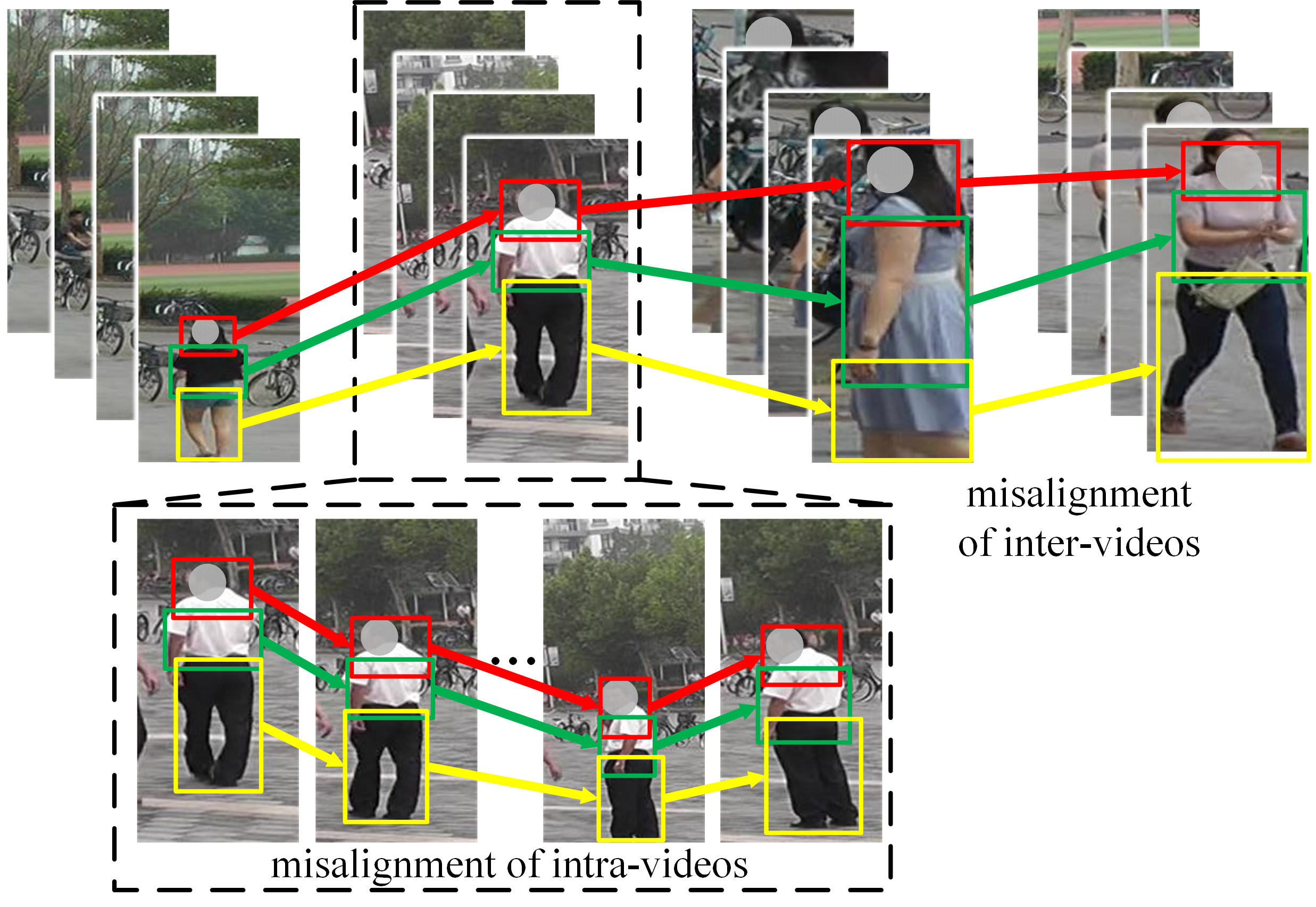}
\caption{The challenge of human region misalignment in video-based person re-ID.}
\label{fig1}
\end{figure}

Video-based re-ID task needs to aggregate features from multiple frames in video sequence. Some video-based person re-ID methods focus on global feature extraction and generating sequence-level features through traditional average pooling or spatial and temporal attention module \cite{6, 7, 8, 9}. In some cases, extracting a global representation from the whole sequence may overlook local details. In order to mine local discriminative features, some methods divide global images into several body regions and learn global and local features simultaneously \cite{1, 2, 5, 10}. However, because of the inaccurate person detections and pose changes, most part-based methods suffer from the problem of region misalignment of both intra- and inter-videos, as illustrated in Fig. \ref{fig1}. 
To align parts, some part-based methods take human parsing or semantic attributes into consideration and they will cost much more computation when a series of frames need to be preprocessed \cite{3, 11}. To effectively deal with part misalignment and avoid excessive computation, we propose a \textbf{Reference-Aided Part-Aligned} (\textbf{RAPA}) framework for video-based person re-ID.

Our proposed RAPA framework is motivated by the application of reference-aided feature learning strategy \cite{5} as well as the success of relation feature learning strategy \cite{12, 13}. The architecture of our method mainly consists of a global feature extracting module, a reference feature learning module and a part feature disentangling module. Specifically, the global feature extracting module, which utilizes the global average pooling and temporal attention block, is applied to extract sequence-level features from the global point of view. The reference feature learning module is developed to find better reference frames and extract discriminative reference features. The part feature disentangling module is deployed to disentangle local features through aligning body parts of video sequences according to references.

We summarize the main contributions of our work into four aspects. First, we propose a novel \textbf{R}eference-\textbf{A}ided \textbf{P}art-\textbf{A}ligned (\textbf{RAPA}) framework for video based person re-ID, which aims to disentangle the discriminative features of different parts. Second, we develop a pose-based \textbf{R}eference \textbf{F}eature \textbf{L}earning (\textbf{RFL}) module to provide the uniform standard for alignment. Several discriminative reference features are extracted from reference frames to ensure the accurate alignment between different videos. Third, we design a relation-based \textbf{P}art \textbf{F}eature \textbf{D}isentangling (\textbf{PFD}) module, which aligns the body parts of intra-video. In this module, a relation-based attention block is adopted to search for corresponding body parts across frames. Finally, we evaluate the performance of the proposed RAPA framework on three mainstream benchmarks: MARS, iLIDS-VID and PRID-2011. Comprehensive experiments display show the superiority of our method to the state-of-the-arts.

\section{Related Work}

\textbf{Video-based Person Re-identification.} Compared with image-based person re-ID, due to the more practical application prospects and much richer spatial-temporal information, video-based person re-ID has achieved more and more attention. Recurrent neural network (RNN) is a common network widely applied to analyze video sequence data. \cite{6} and \cite{8} introduced the models combining CNN and RNN to extract frame-level features and aggregate them by temporal pooling. However, these models treat all frames equally so that poor-quality frames and extraneous spatial regions influence the discriminability of features. To make the network focus on more informative frames and regions, attention mechanism is applied in video-based person re-identification. \cite{9} proposed temporal attention models to calculate weight scores for time steps. Furthermore, \cite{4} and \cite{14} adopted attention mechanism in both spatial and temporal dimension. Compared with these existing video-based attention methods, our attention module in the proposed RAPA framework needs no additional parameters and mines attention scores according to the relations between reference features and frame-level feature maps.

\textbf{Part-based Person Re-identification.} In order to mine local informative details, some recent works divided pedestrian images into several parts and focused on local discriminative feature learning. One of the most intuitive and concise partition strategies is hard segmentation \cite{1, 2}. \cite{1} introduced a part-based convolutional baseline which divided pedestrian images vertically and equidistantly. \cite{2} adopted a multiple granularities partition strategy to capture features of different scales. Compared with the hard segmentation, segmentation based on human pose and semantic information can avoid the problem of misalignment of body parts \cite{3, 11}. However, most of traditional pose-based and semantics-based methods are not applicable to video-based person re-ID due to the huge and complex computation when they preprocess all frames of videos. Instead, our method only segments the reference frame according to pose information and utilizes references to align human parts.

\textbf{Reference-aided Feature Learning Strategy.} Owing to the continuity of video sequence, no significant difference exists in appearance or structure between consecutive frames. Accordingly, a single frame can become a reference and provide guidance for the analysis of the whole video sequence. Some recent works \cite{5, 15} have applied reference-aided feature learning strategy in person re-ID. In \cite{5}, the first frame of each video is processed to guide subsequent frames, while in \cite{15}, the average of a video sequence is regarded as the reference frame. However, some poor-quality frames cannot provide enough reference value for video sequences. The quality evaluation block in our proposed RFL module solves this problem, which can select high quality frames automatically to generate better references.

\section{Proposed Method}

\subsection{Framework Overview}
\begin{figure}[t]
\centering
\includegraphics[width=1.0\columnwidth]{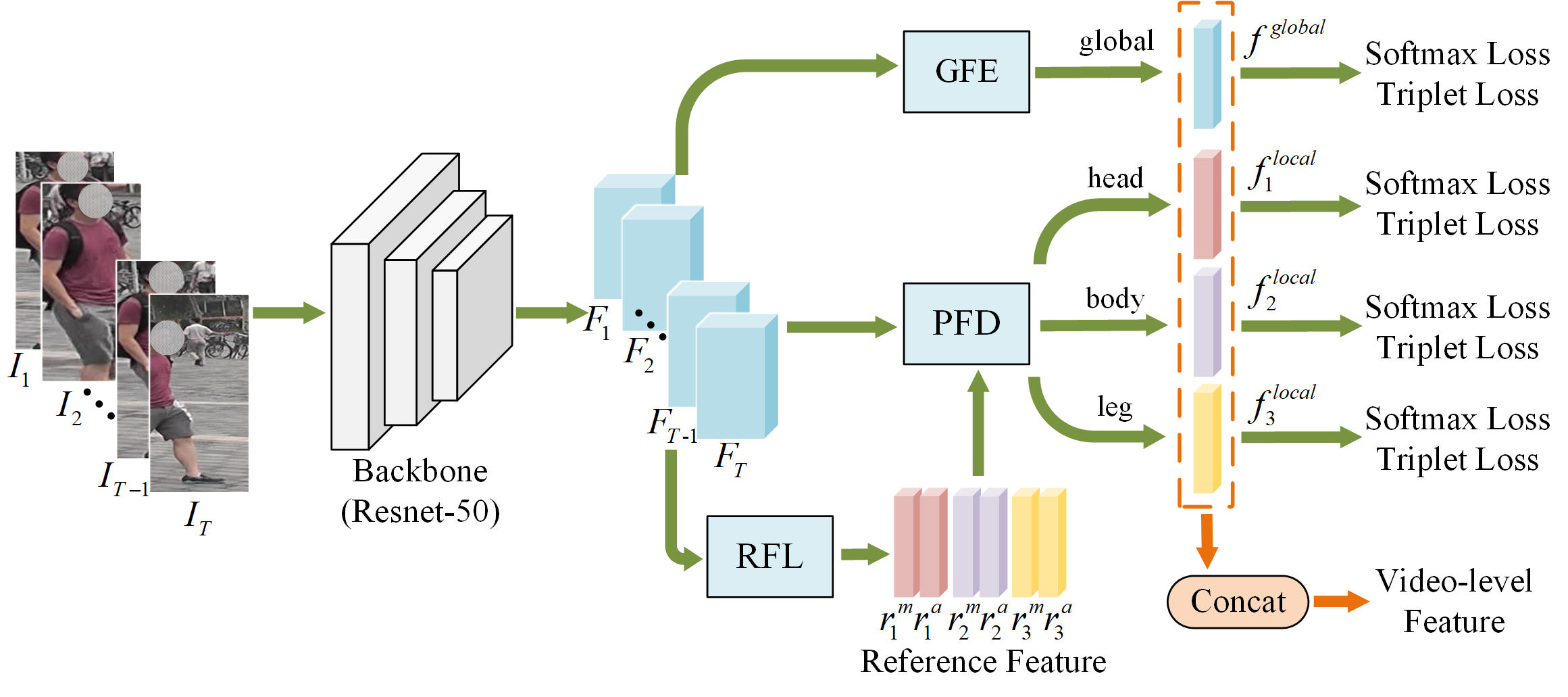}
\caption{The architecture of the proposed RAPA. This framework mainly contains three modules, including global feature extracting (GFE) module, reference feature learning (RFL) module and part feature disentangling (PFD) module. The GFE module is applied for the global spatiotemporal feature extraction, and RFL and PFD modules are jointly deployed to extract several local features.}
\label{fig2}
\end{figure}

The overview of our proposed framework is shown in Fig. \ref{fig2}. We select $T$ frames from a video sequence as $V{\rm{ = }}\left\{ {{I_t}} \right\}_{t = 1}^T$, and the feature map of each frame ${F_t} \in {{\rm{\mathbb{R}}}^{C \times H \times W}}$ is extracted through the backbone (e.g., ResNet-50), where $C$, $H$ and $W$ represent the channel, height and width of the feature maps, respectively. 
In the global branch, the feature maps are fed into Global Feature Extracting (GFE) module to extract features from the global point of view. Following the methods proposed in \cite{5, 7}, the feature maps are aggregated into image-level representations by the global average pooling, and then they are fused to video-level representations through temporal attention mechanism. After that, we use a $1 \times 1$  convolutional layer to reduce the dimension of features and get final global features denoted as ${f^{global}} \in {{\rm{\mathbb{R}}}^{\frac{C}{s}}}$, where $s$ controls the dimension reduction ratio.

The local branch mainly contains Reference Feature Learning (RFL) module and Part Feature Disentangling (PFD) module. The former is used to find better local reference features, which can provide guidance for the alignment of video sequences, and the latter is used to align part features according to the references. Three local features (head, body and leg parts) extracted from both mentioned modules are denoted as $f_p^{local} \in {{\rm{\mathbb{R}}}^{\frac{C}{s}}}$ ($p \in \left[ {1,3} \right]$). These two modules will be explained in detail in following subsections. The final video-based representation $f \in {{\rm{\mathbb{R}}}^{4 \times \frac{C}{s}}}$ can be obtained by concatenating the global and local features:

\begin{equation}
    f = \left[ {{f^{global}},f_1^{local},f_2^{local},f_3^{local}} \right]
\end{equation}

\subsection{Pose-based Reference Feature Learning Module}

\begin{figure}[t]
\centering
\includegraphics[width=1.0\columnwidth]{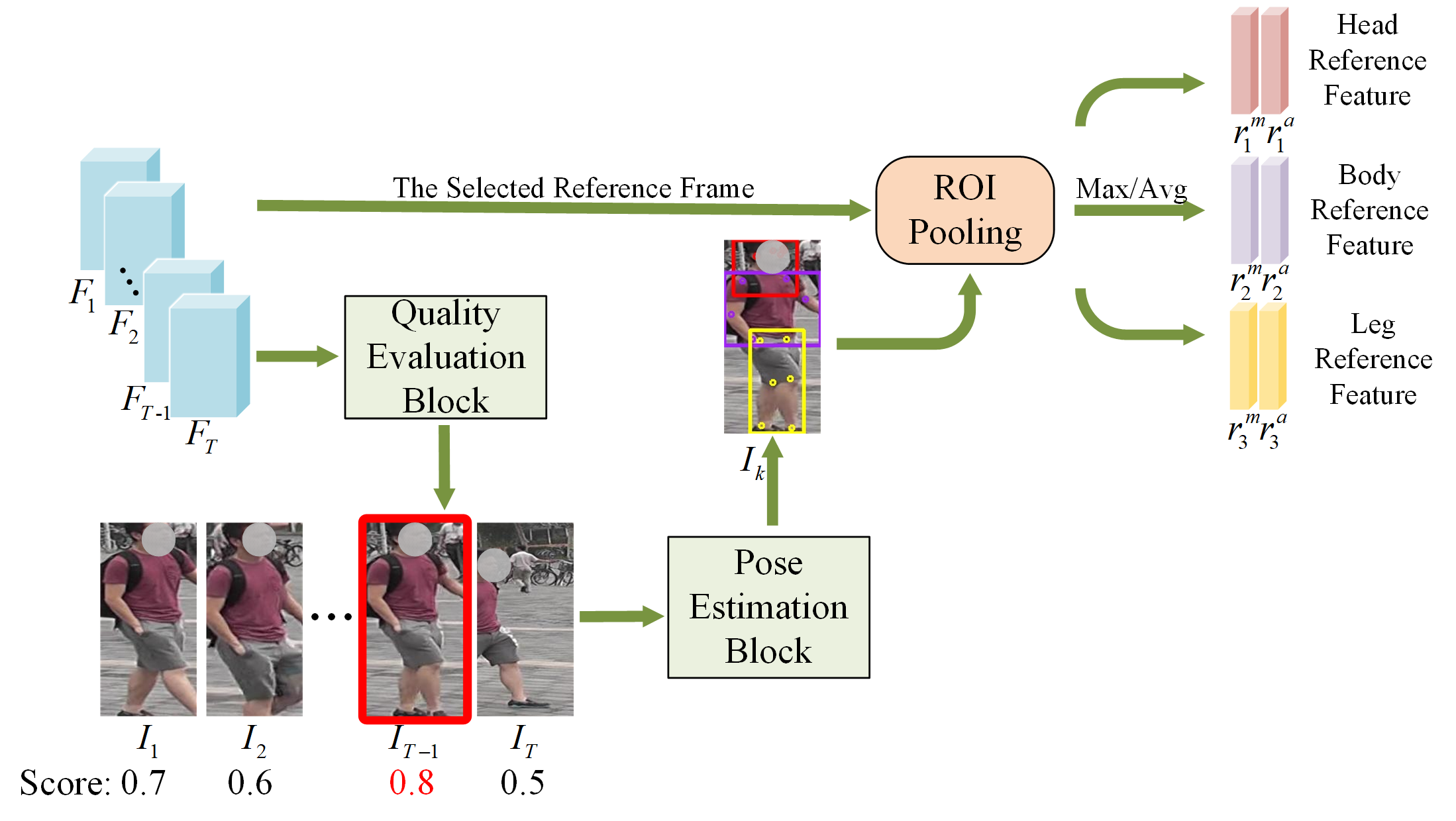}
\caption{The details of the pose-based Reference Feature Learning (RFL) module.}
\label{fig3}
\end{figure}

In \cite{5}, the first frame of the input video sequence is taken as a reference frame, which may be not in good condition due to the inaccurate person detections and occlusions.
The quality of reference features determines the effect of alignment between videos.  Therefore, we develop a posed-based reference feature learning (RFL) module to generate high-quality reference features and align part features between different videos, as illustrated in Fig. \ref{fig3}.

Firstly, in order to estimate the quality of images and find a better reference frame, we design a quality evaluation block which is motivated by temporal attention. Given the feature maps ${F_t} \in {\mathbb{R}^{C \times H \times W}}$ ($t \in \left[ {1,T} \right]$), we get the image-level feature vectors ${l_t} \in {\mathbb{R}^C}$ with a global average pooling layer. The quality scores of frames are calculated by:
\begin{equation}
    {q_t} = {\rm{Sigmoid}}\left( {{\rm{BN}}\left( {{\rm{Conv}}\left( {{l_t}} \right)} \right)} \right)
\end{equation}
where the convolutional layer (Conv) reduces the vector dimension to 1 followed by a batch normalization layer (BN) and a sigmoid activation function (Sigmoid). The reference frame in each video sequence is defined as the frame which obtains the maximum quality score, denoted as ${I_k}$ where $k = \arg \max \left( {{q_t}} \right)$.

Secondly, we apply the human pose estimation model (e.g., HRNet \cite{16}) on ${I_k}$ to predict human key-points. According to the distribution of key-points, the human body is divided into three parts including head, body and leg. The ROI pooling is commonly applied to capture the regions of interest from the whole feature map, which is widely used in object detection. In RFL module, both max and average ROI pooling are used to extract local reference features, because the former can focus on the image texture and the latter can ensure the integrity of information. We denote the local reference features as $r_1^m$, $r_1^a$, $r_2^m$, $r_2^a$, $r_3^m$ and $r_3^a \in {{\rm{\mathbb{R}}}^C}$ ($m$ means max pooling, $a$ means average pooling, and 1-3 means head, body and leg), respectively.

\begin{figure}[t]
\centering
\includegraphics[width=1.0\columnwidth]{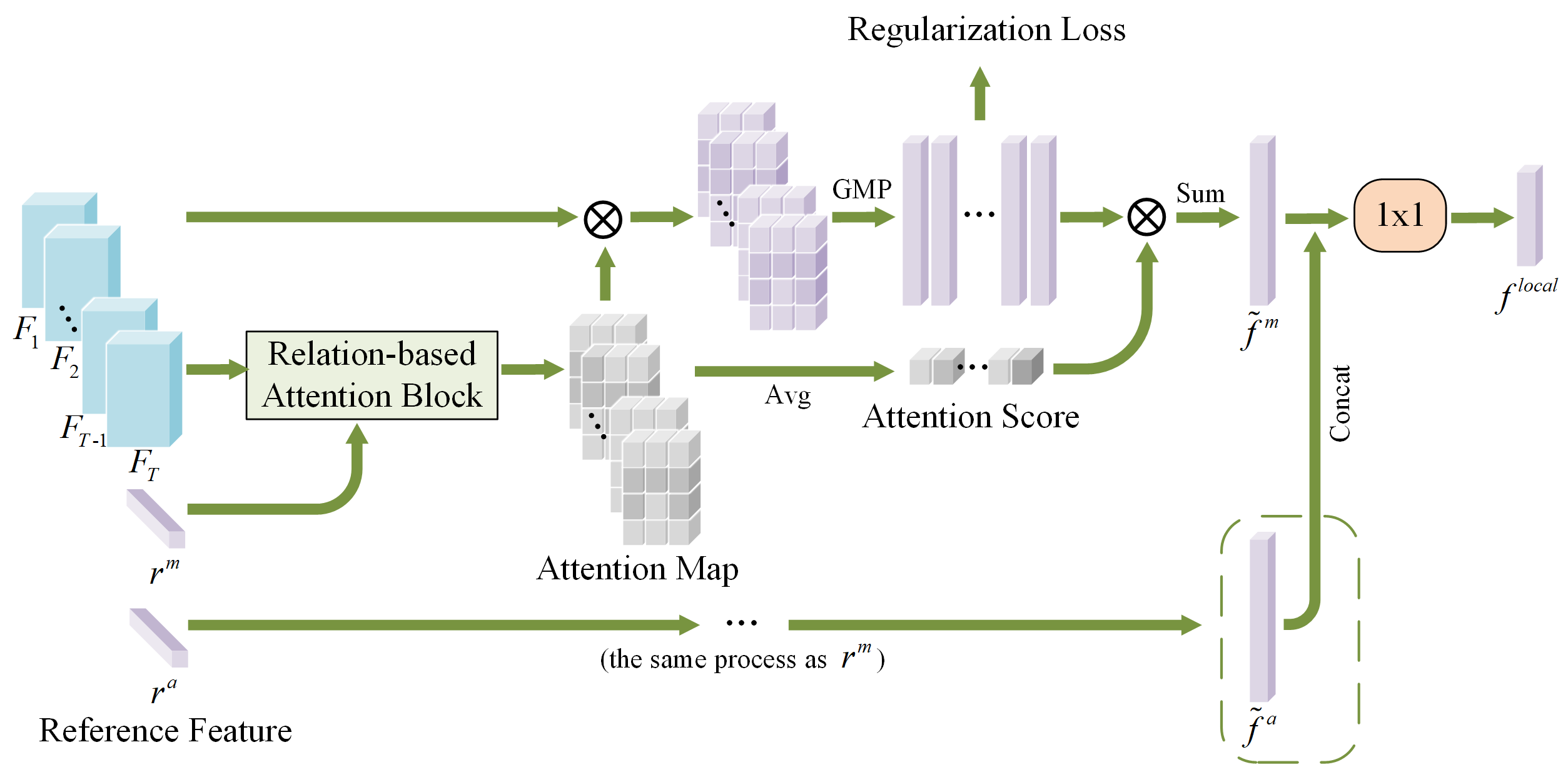}
\caption{The details of the relation-based Part Feature Disentangling (PFD) module.}
\label{fig4}
\end{figure}

\subsection{Relation-based Part Feature Disentangling Module}

The reference features provide the guidance for the alignment of intra-video sequences. To precisely disentangle local features, we introduce a relation-based part feature disentangling (PFD) module as shown in Fig. \ref{fig4}. In this paper, the vector along the channel dimension in feature maps is denoted as a column vector. We can measure the relevance between each reference feature vector and each column vector to obtain local attention maps in the relation-based attention block. Given the reference feature vectors $r_p^m$, $r_p^a$ ($p \in \left[ {1,3} \right]$) and column vectors $v_t^{h,w} \in {{\rm{\mathbb{R}}}^C}$ in feature maps ${F_t}$ ($t \in \left[ {1,T} \right]$, $h \in \left[ {1,H} \right]$, $w \in \left[ {1,W} \right]$), each relation element in relation map $D_{p,t}^m \in {{\rm{\mathbb{R}}}^{C \times H \times W}}$ is calculated by: 
\begin{equation}
    d_{p,t,h,w}^m = {\left( {v_t^{h,w} - r_p^m} \right)^2}
\end{equation}
After that, a batch normalization layer and a sigmoid activation function are applied to normalize each relation element to the range of $\left( {0,1} \right)$ and obtain the attention map $A_{p,t}^m \in {{\rm{\mathbb{R}}}^{C \times H \times W}}$ by:
\begin{equation}
    A_{p,t}^m = E{\rm{ - Sigmoid}}\left( {{\rm{BN}}\left( {D_{p,t}^m} \right)} \right)
\end{equation}
where $E \in {{\rm{\mathbb{R}}}^{C \times H \times W}}$ is a matrix in which all elements are 1. Through the spatial attention mechanism and global max pooling (GMP), the elements in feature maps with high relevance can be found and aggregated to image-level local features $f_{p,t}^m \in {{\rm{\mathbb{R}}}^C}$ as formulated:
\begin{equation}
    f_{p,t}^m = {\rm{GMP}}\left( {{F_t} * A_{p,t}^m} \right)
\end{equation}
where $*$ is Hadamard product. Besides, in order to promote this module to focus on the more informative frames, the temporal channel attention mechanism is applied to weight the image-level local features. Based on the attention map, the attention score $S_{p,t}^m \in {{\rm{\mathbb{R}}}^C}$ is computed as: 
\begin{equation}
    S_{p,t}^m = \frac{1}{{H \times W}}A_{p,t}^m
\end{equation}
Then we can get the video-level aligned local feature $\tilde f_p^m \in {{\rm{R}}^C}$ through weighted sum:
\begin{equation}
    \tilde f_p^m = \sum\limits_{t = 1}^T {f_{p,t}^m * S_{p,t}^m}
\end{equation}
where $*$ is Hadamard product. The calculation of $\tilde f_p^a$ is similar to $\tilde f_p^m$, and we omit the description of this part for convenience. Finally, we concatenate these two aligned features as $\left[ {\tilde f_p^m,\tilde f_p^a} \right]$ and get the final part feature $f_p^{local} \in {{\rm{\mathbb{R}}}^{\frac{C}{s}}}$ by performing a $1 \times 1$ convolutional layer on it to reduce its dimension, where $s$ controls the dimension reduction ratio.

Due to the similarity of continuous frames, we design the inter-frame regularization term to promote the relation-based attention block to maintain the similarity between attention maps of different frames and avoid focusing on only one frame. Specifically, the regularization term of each video sequence is:
\begin{equation}
    Reg = \sum\limits_{i = 1}^T {\sum\limits_{j = 1,j \ne i}^T {\sum\limits_{p = 1}^3 {\left( {\left\| {f_{p,i}^m - f_{p,j}^m} \right\|_2^2 + \left\| {f_{p,i}^a - f_{p,j}^a} \right\|_2^2} \right)} } }
\end{equation}

\subsection{Loss Function}
In our framework, we adopt both batch hard triplet loss and softmax cross entropy loss on each mini branch, as shown in Fig. \ref{fig2}. We assume that each mini-batch consists of $P$ identities and $K$ tracklets of each identity. The triplet loss for each branch is calculated by:
\begin{equation}
\begin{array}{r}
{L_{tri}} = \sum\limits_{i = 1}^P {\sum\limits_{a = 1}^K {[m + \overbrace {\mathop {\max }\limits_{p = 1...K} {{\left\| {f_a^{\left( i \right)} - f_p^{\left( i \right)}} \right\|}_2}}^{{\rm{hardest\ positive}}}} } \\
 - \underbrace {\mathop {\min }\limits_{\scriptstyle n = 1...K\atop
{\scriptstyle j = 1...P\atop
\scriptstyle j \ne i}} {{\left\| {f_a^{\left( i \right)} - f_n^{\left( j \right)}} \right\|}_2}}_{{\rm{hardest\ negative}}}{]_ + }
\end{array}
\end{equation}
where $f_a^{\left( i \right)}$, $f_p^{\left( i \right)}$ and $f_n^{\left( j \right)}$ are the features extracted from the anchor, positive and negative samples respectively, and $m$ is the margin hyperparameter to control the differences between intra and inter distances. The softmax cross entropy loss for each branch is formulated as:
\begin{equation}
    {L_{softmax}} =  - \frac{1}{{P \times K}}\sum\limits_{i = 1}^P {\sum\limits_{a = 1}^K {{y_{i,a}}\log {q_{i,a}}} }
\end{equation}
where ${y_{i,a}}$ and ${q_{i,a}}$ are the ground truth identity and prediction of tracklet sample $\left\{ {i,a} \right\}$. Therefore, the total loss for each branch is:
\begin{equation}
L_c^{branch} = {L_{tri}} + {L_{softmax}}
\end{equation}
where $c \in \left[ {1,4} \right]$ indicates the branch number and $L_1^{branch}$ and $\left\{ {L_c^{branch}} \right\}_{c = 2}^4$ indicate the loss from global branch and three local branches, respectively. Besides, the inter-frame regularization loss can be calculated by:
\begin{equation}
    {L_{reg}} = \frac{1}{{P \times K}}\sum\limits_{}^P {\sum\limits_{}^K {Reg} }
\end{equation}
The addition of branch losses and regularization loss constitutes the final loss for optimization:
\begin{equation}
    {L_{total}} = \sum\limits_{c = 1}^4 {L_c^{branch} + \lambda {L_{reg}}}
\end{equation}
where $\lambda$ is a hyper-parameter to control the proportion of regularization loss.

\section{Experiments}
\subsection{Datasets and Evaluation Protocol}

\textbf{Datasets.} Three standard video-based person re-ID datasets: iLIDS-VID \cite{17}, PRID-2011 \cite{18} and MARS \cite{19}, are applied to evaluate our proposed framework. \textbf{iLIDS-VID} dataset, which is challenging due to occlusion and blur, consists of 300 identities and each identity includes 2 video sequences taken from a pair of non-overlapping cameras. By comparsion, \textbf{PRID-2011} dataset is less challenging due to its simple environments with rare occlusion. It contains 385 and 749 identities from 2 different cameras, but only the first 200 identities appear in both cameras. \textbf{MARS} dataset is one of the largest video-based person re-ID benchmarks, which consists of 1261 identities and 20,715 video sequences captured by 6 cameras.

\textbf{Evaluation Protocol.} The CMC curve and mAP are applied to evaluate the performance of our framework. For iLIDS-VID and PRID-2011 datasets, following the common practices in previous work \cite{17}, we randomly split the half-half for training and testing. The average result of 10 repeated experiments is reported. For MARS dataset, 8298 sequences of 625 identities are used for training and other sequences are used for testing.

\begin{figure}[t]
\centering
\includegraphics[width=0.9\columnwidth]{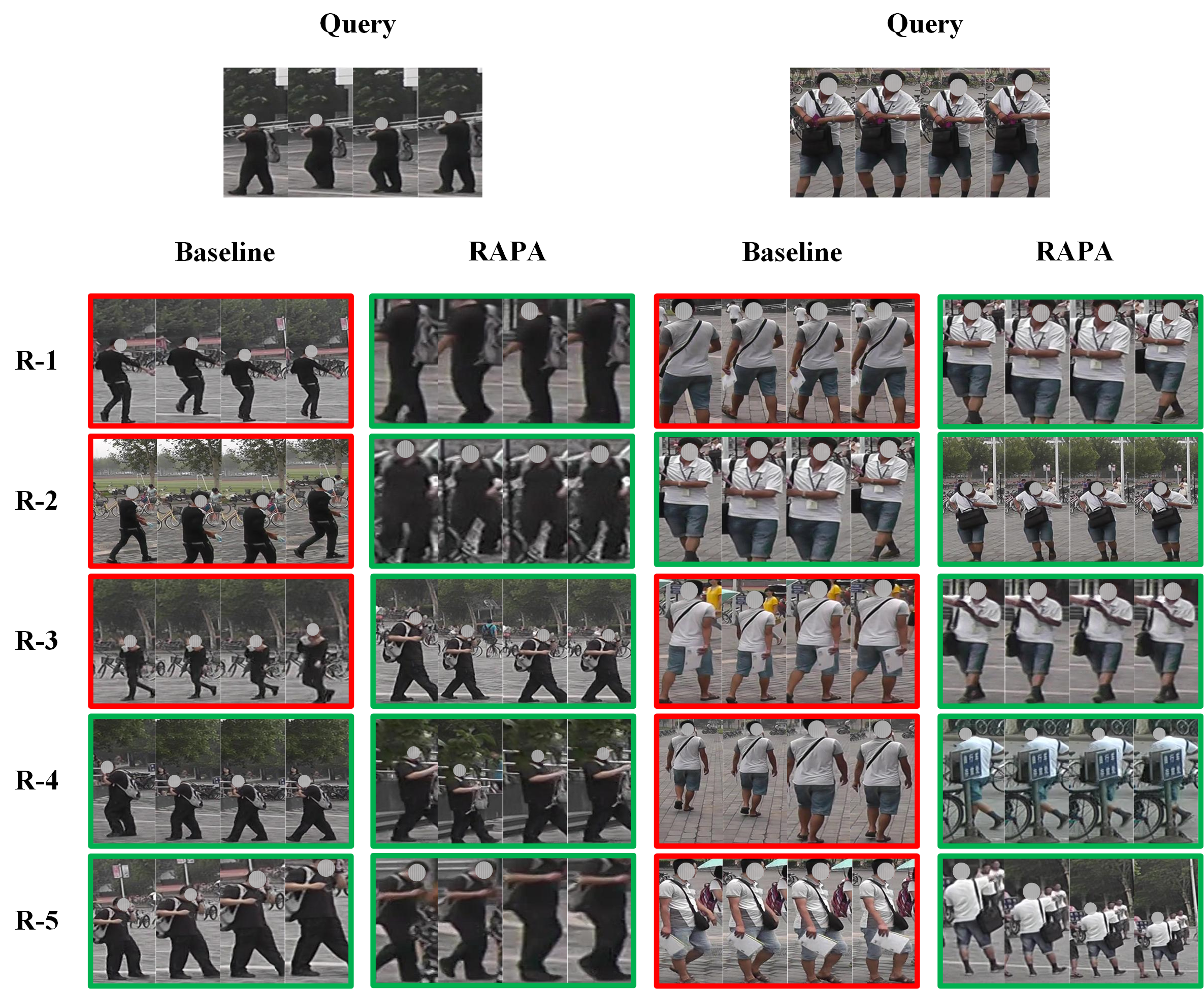}
\caption{Visualization of person re-ID results using the baseline model and our proposed RAPA framework. The green and red bounding boxes indicate correct and incorrect matches, respectively.}
\label{fig6}
\end{figure}

\begin{table}[t]
\centering
\caption{Ablation study on the components of our proposed method on MARS dataset.}
\resizebox{\columnwidth}{!}{
\begin{tabular}{l c c c c}
    \hline
    Variant & Rank-1 & Rank-5 & Rank-20 & mAP \\
    \hline
    (a) Baseline & 82.4 & 93.8 & 97.1 & 74.1 \\
    (b) Baseline+RA & 84.7 & 95.1 & 98.1 & 76.3 \\
    (c) Baseline+RA+Reg & 86.7 & 96.2 & 98.1 & 81.0 \\
    (d) Baseline+RA+Reg+TA & 87.7 & 96.1 & 98.2 & 82.2 \\
    (e) Baseline+RA+Reg+TA+PE & 88.0 & \textbf{96.5} & \textbf{98.2} & 82.7 \\
    (f) Baseline+RA+Reg+TA+PE+QE & \textbf{88.7} & 96.1 & 98.1 & \textbf{82.8} \\
    \hline
\end{tabular}}
\label{tab1}
\end{table}

\subsection{Implementation Details}
In the training phase, we randomly select $T = 4$ frames of each video as the input sequence. Input images are resized to $128 \times 256$. Random erasing and random cropping are applied for data augmentation. A mini-batch consists of 8 identities with 4 tracklets. The ResNet-50 pretrained on ImageNet is utilized as our backbone. We choose Adam to optimize our model with weight decay $5 \times {10^{ - 4}}$. The initial learning rate is $3.5 \times {10^{ - 4}}$ and decreased by 0.1 every 100 epochs. The epoch number is 400 in total. For iLIDS-VID and PRID-2011 datasets, the hyper-parameter $\lambda$ in final loss function is set to $5 \times {10^{ - 5}}$, and for MARS dataset, $\lambda$ is set to $3 \times {10^{ - 4}}$. In the testing phase, the video sequence is segmented into several clips of length $T = 4$. The average of all clip-level features of the same sequence is regarded as the video-level feature. Euclidean distance is applied to measure the similarity between query sequences and gallery sequences.

\subsection{Ablation Study}

To evaluate the effectiveness of components in our proposed RAPA framework, we conduct a series of ablation experiments and show the comparative results in Table \ref{tab1}. We select the global branch without $1 \times 1$ convolutional layer as our baseline, which follows the method proposed in \cite{7}. The PFD module is divided into \textbf{RA}, \textbf{Reg} and \textbf{TA}, which correspond to the relation-based attention block, the inter-frame regularization loss and the temporal channel attention score, respectively. The RFL module includes \textbf{PE} and \textbf{QE}, which indicate the pose estimation block and the quality evaluation block, respectively. Compared with the baseline, our framework improves Rank-1 and mAP from 82.4\% and 74.1\% to \textbf{88.7\%} and \textbf{82.8\%} on MARS dataset. Some comparative results are visualized in Fig. \ref{fig6}. As can be observed intuitively, misalignments bring difficulties for the baseline model to distinguish some pedestrians with similar appearances, while our proposed RAPA framework can achieve more robust results in these cases.

\textbf{Effectiveness of PFD Module.} The comparative results of variants (a) – (f) show the effectiveness of our proposed PFD module. Variant (a) doesn’t perform well because it only takes the global feature into consideration but ignores the more discriminative local details. Variant (b) utilizes the relation-based attention block which can focus on the corresponding local areas. Without the RFL module, variant (b) applies the hard segmentation on the first frame to generate the reference features by default. It can be observed that \textbf{RA} improves Rank-1 and mAP accuracy by 2.3\% and 2.2\%, respectively. The application of \textbf{Reg} in variant (c) preserves the similarity between continuous frames and further improves the accuracy of our framework. To encourage the framework to focus on the frames of interest, \textbf{TA} is adopted in variant (d) which forms a complete PFD module. In summary, compared with the baseline, our framework with PFD module achieves 5.3\% and 8.1\% improvements in Rank-1 and mAP, respectively.

\begin{table}[t]
\centering
\caption{Ablation study on the branches of feature representation on MARS dataset.}
\resizebox{\columnwidth}{!}{
\begin{tabular}{l c c c c}
    \hline
    Variant & Rank-1 & Rank-5 & Rank-20 & mAP \\
    \hline
    (a) Global (Baseline) & 82.4 & 93.8 & 97.1 & 74.1 \\
    (b) Local & 87.4 & \textbf{96.6} & \textbf{98.3} & 81.7 \\
    (c) Global+Local & \textbf{88.7} & 96.1 & 98.1 & \textbf{82.8} \\
    \hline
\end{tabular}}
\label{tab2}
\end{table}

\begin{table}[t]
\centering
\caption{Comparisons of our proposed method to the state-of-the-art methods on MARS, iLIDS-VID and PRID-2011 datasets. The ${{\rm{1}}^{st}}$, ${{\rm{2}}^{nd}}$ and ${{\rm{3}}^{rd}}$ best results are emphasized with red , blue and green color, respectively.}
\resizebox{\columnwidth}{!}{
\begin{tabular}{l c c c c c c c c c c}
    \hline
    \multirow{2}{*}{Method} & \multirow{2}{*}{Publication} & \multicolumn{2}{c}{MARS} & \multicolumn{1}{c}{iLIDS-VID} & \multicolumn{1}{c}{PRID-2011} \\
    \cline{3-6}
       & & Rank-1 & mAP & Rank-1 & Rank-1 \\
    \hline
    RNN \cite{6} & CVPR'16 & - & - & 58.0 & 70.0 \\
    CNN+XQDA \cite{19} & ECCV'16 & 68.3 & 49.3 & 53.0 & 77.3 \\
    CRF \cite{8} & CVPR'17 & 71.0 & - & 61.0 & 77.0 \\
    SeeForest \cite{9} & CVPR'17 & 70.6 & 50.7 & 55.2 & 79.4 \\
    RQEN \cite{20} & AAAI'18 & 77.8 & 71.1 & 76.1 & 92.4 \\
    STAN \cite{14} & CVPR'18 & 82.3 & 65.8 & 80.2 & 91.2 \\
    STA \cite{4} & AAAI'19 & 86.3 & 80.1 & - & - \\
    RRU \cite{21} & AAAI'19 & 84.4 & 72.7 & 84.3 & 92.7 \\
    A3D \cite{22} & TIP'20 & 86.3 & \color{green}{\textbf{80.4}} & \color{green}{\textbf{87.9}} & 95.1 \\
    AMEM \cite{23} & AAAI'20 & 86.7 & 79.3 & 87.2 & 93.3 \\
    FGRA \cite{5} & AAAI'20 & \color{green}{\textbf{87.3}} & \color{blue}{\textbf{81.2}} & \color{blue}{\textbf{88.0}} & \color{blue}{\textbf{95.5}} \\
    Two-stream M3D \cite{24} & TIP'20 & \color{blue}{\textbf{88.6}} & 79.5 & 86.7 & \color{red}{\textbf{96.6}} \\
    \hline
    Ours & ICME'21 & \color{red}{\textbf{88.7}} & \color{red}{\textbf{82.8}} & \color{red}{\textbf{89.6}} & \color{green}{\textbf{95.2}} \\
    \hline
\end{tabular}}
\label{tab3}
\end{table}

\textbf{Effectiveness of RFL Module}. The comparable results of variants (e) and (f) prove the effectiveness of our proposed RFL module. Variant (e) utilizes \textbf{PE} block to segment frames into several parts according to the pose information. Compared with variant (d), variant (e) solves the misalignment between different video sequences. Variant (f) further deploys \textbf{QE} block to find high-quality reference frames which improves the robustness in some complex cases such as occlusion. Finally, variant (f) can outperform variant (d) by 1\% and 0.6\% in Rank-1 and mAP on MARS dataset.

\textbf{Effectiveness of Different Branches.} As shown in Fig. \ref{fig2}, the feature representation in our framework consists of global branch and 3 local branches. To verify the effectiveness of these branches, several ablation experiments are conducted and the comparable results are shown in Table~\ref{tab2}. Variant (a) only contains the global branch which is our baseline. Variant (b) disentangles several local features and achieves a significant improvement. Compared with variant (b), variant (c) achieves the best results owing to the information compensation from global branch.

\subsection{Comparison with the State-of-the-arts}
Table~\ref{tab3} reports the results of our proposed RAPA framework and some state-of-the-art methods on MARS, iLIDS-VID and PRID-2011 datasets. As we can observe from Table~\ref{tab3}, our model obtains the Rank-1 accuracy of \textbf{88.7\%}, \textbf{89.6\%} and \textbf{95.2\%} on MARS, iLIDS-VID and PRID-2011, which outperforms all the state-of-the-art methods in most evaluation indicators. The main reason is that our framework accomplishes the feature alignment of both intra- and inter-video sequences. However, our RAPA framework has the slightly lower accuracy than FGRA and Two-stream M3D on PRID-2011 dataset. One possible explanation is that the pedestrian images in PRID-2011 are perfectly neat and the condition of misalignment is rare, so that the alignment strategy cannot exert its advantage.

\section{Conclusion}
In this paper, a reference-aided part-aligned (RAPA) framework is proposed for video-based person re-identification. In order to solve the pedestrian misalignment problem, the pose-based reference feature learning (RFL) module and the relation-based part feature disentangling (PFD) module are explored. The former is designed to extract discriminative reference features and align the local features between different video sequences. The latter is applied to align parts of intra-video sequences according to the references. Moreover, in PFD module, a relation-based attention block is adopted to search for corresponding body parts. The outstanding experimental results prove the superiority of the proposed RAPA framework.

\small
\bibliographystyle{IEEEbib}


\end{document}